\documentclass[conference]{IEEEtran}
\IEEEoverridecommandlockouts

\IEEEoverridecommandlockouts
\usepackage{cite}
\usepackage{amsmath,amssymb,amsfonts}
\usepackage{algorithmic, amssymb, gensymb}
\usepackage{graphicx}
\usepackage{textcomp}
\usepackage{xcolor}
\usepackage{multirow}
\usepackage{float}

\def\BibTeX{{\rm B\kern-.05em{\sc i\kern-.025em b}\kern-.08em
    T\kern-.1667em\lower.7ex\hbox{E}\kern-.125emX}}

\usepackage[caption=false]{subfig}

\title{\LARGE \bf
Object Pose Distribution Estimation for Determining Revolution and Reflection Uncertainty in Point Clouds
\thanks{This project was funded in part by Innovation Fund Denmark through the projects MADE ReAct and FERA, and in part by the SDU I4.0-Lab.}
}

\author{\IEEEauthorblockN{1\textsuperscript{st} Frederik Hagelskjær*}
\IEEEauthorblockA{\textit{SDU Robotics} \\
\textit{The Mærsk Mc-Kinney} \\ 
\textit{Møller Institute} \\
\textit{University of Southern Denmark}\\
Odense, Denmark \\
frhag@mmmi.sdu.dk}
*Corresponding author
~\\
\and
\IEEEauthorblockN{2\textsuperscript{nd} Dimitrios Arapis}
\IEEEauthorblockA{
\textit{Finished Product} \\
\textit{Manufacturing} \\
\textit{Science \& Technology} \\ % Finished Product Manufacturing
\textit{Novo Nordisk A/S}\\
Hillerod, Denmark \\
dtai@novonordisk.com}
~\\
\and
\IEEEauthorblockN{3\textsuperscript{rd} Steffen Madsen}
\IEEEauthorblockA{
\textit{Finished Product} \\
\textit{Manufacturing} \\
\textit{Science \& Technology} \\ % Finished Product Manufacturing
\textit{Novo Nordisk A/S}\\
Hillerod, Denmark \\
bfma@novonordisk.com}
~\\
\and
\IEEEauthorblockN{4\textsuperscript{th} Thorbjørn Mosekjær Iversen}
\IEEEauthorblockA{\textit{SDU Robotics} \\
\textit{The Mærsk Mc-Kinney} \\ 
\textit{Møller Institute} \\
\textit{University of Southern Denmark}\\
Odense, Denmark \\
thmi@mmmi.sdu.dk}
}

\begin{document}
\maketitle

\begin{abstract}
Object pose estimation is crucial to robotic perception and typically provides a single-pose estimate. However, a single estimate cannot capture pose uncertainty deriving from visual ambiguity, which can lead to unreliable behavior. Existing pose distribution methods rely heavily on color information, often unavailable in industrial settings.

We propose a novel neural network-based method for estimating object pose uncertainty using only 3D colorless data. To the best of our knowledge, this is the first approach that leverages deep learning for pose distribution estimation without relying on RGB input. We validate our method in a real-world bin picking scenario with objects of varying geometric ambiguity. Our current implementation focuses on symmetries in reflection and revolution, but the framework is extendable to full SE(3) pose distribution estimation. Source code available at opde3d.github.io
\end{abstract}

\begin{figure}[ht]
\centering
% \subfloat[View where the square recess is visible. (Red indent at top).]{\includegraphics[trim={6.5cm 1.5cm 6.5cm 3cm},clip, width=.48\linewidth]{gfx/Screenshot from 2025-07-29 13-51-29_2.png}}\
% \subfloat[Rotated view without any distinguishing features.]{\includegraphics[trim={6.5cm 1.5cm 6.5cm 3.0cm},clip, width=.48\linewidth]{gfx/Screenshot from 2025-07-29 13-52-24_2.png}} 
\subfloat{\includegraphics[trim={6.5cm 1.0cm 6.5cm 2cm},clip, width=.46\linewidth]{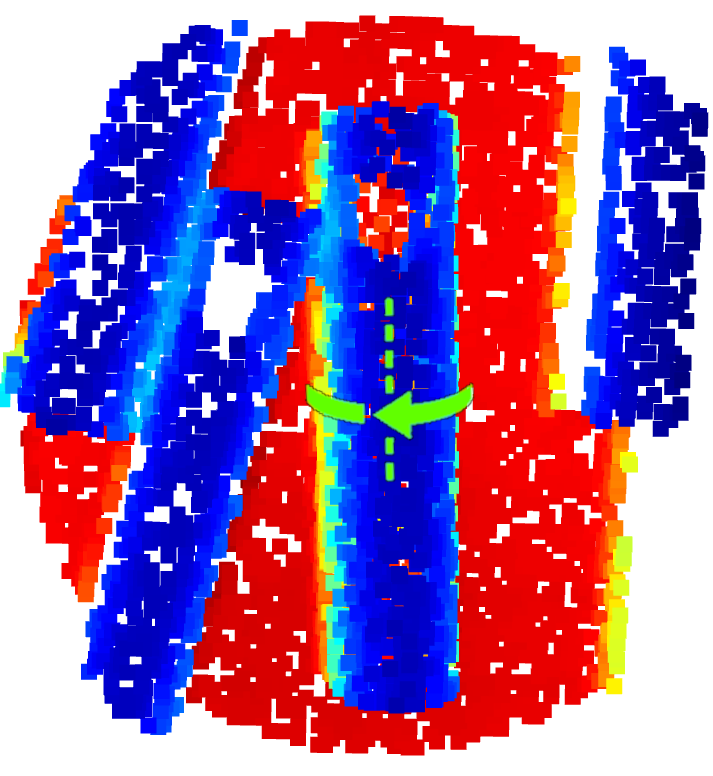}}\
\subfloat{\includegraphics[trim={6.5cm 1.0cm 6.5cm 2.0cm},clip, width=.46\linewidth]{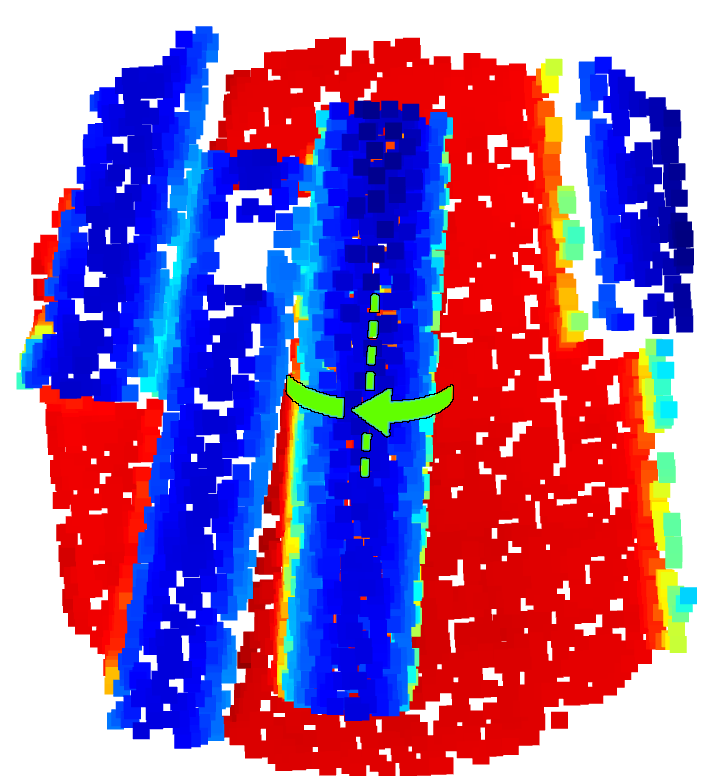}} 
% \subfloat[Unambiguous view with a single object pose.]{\includegraphics[width=.48\linewidth]{gfx/Screenshot from 2025-07-29 13-51-29.png}}\
% \subfloat[Ambiguous view with unknown revolution and rotoreflection.]{\includegraphics[width=.48\linewidth]{gfx/Screenshot from 2025-07-29 13-52-24.png}} 
    \vspace{-1.5mm}
    % \subfloat[Pose distribution for image (a) with a single prediction.]{\includegraphics[trim={0cm 0 1.5cm 0},clip, width=.48\linewidth]{gfx/Figure_1.png}}\
    % \subfloat[Pose distribution for image (b) with large uncertainty in reflection.]{\includegraphics[trim={0cm 0 1.5cm 0},clip,width=.48\linewidth]{gfx/Figure_2.png}} 
    \subfloat[View where the square recess is visible. (Red indent at top). Resulting in a pose distribution with a single prediction.]{\includegraphics[trim={0cm 0cm 1.5cm 1cm},clip, width=.48\linewidth]{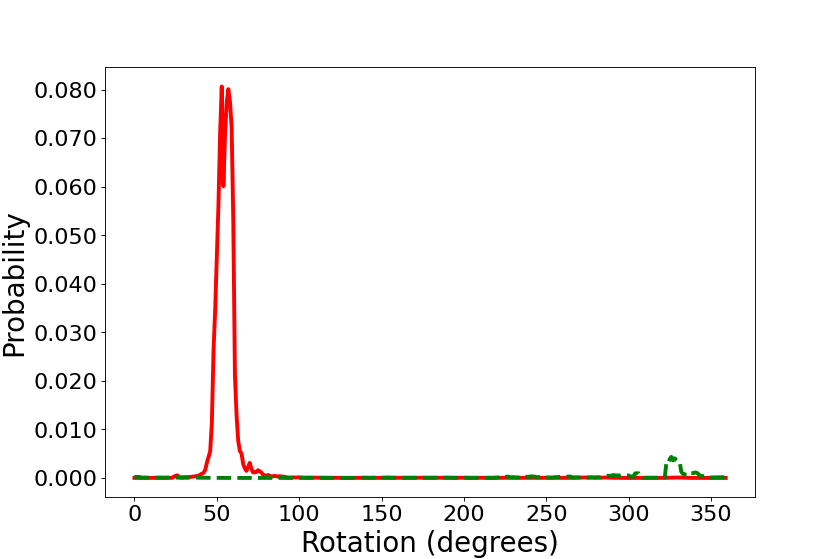}}\
    \subfloat[Rotated view without any distinguishing features. The pose distribution is larger with uncertainty in reflection.]{\includegraphics[trim={0cm 0 1.5cm 1cm},clip,width=.48\linewidth]{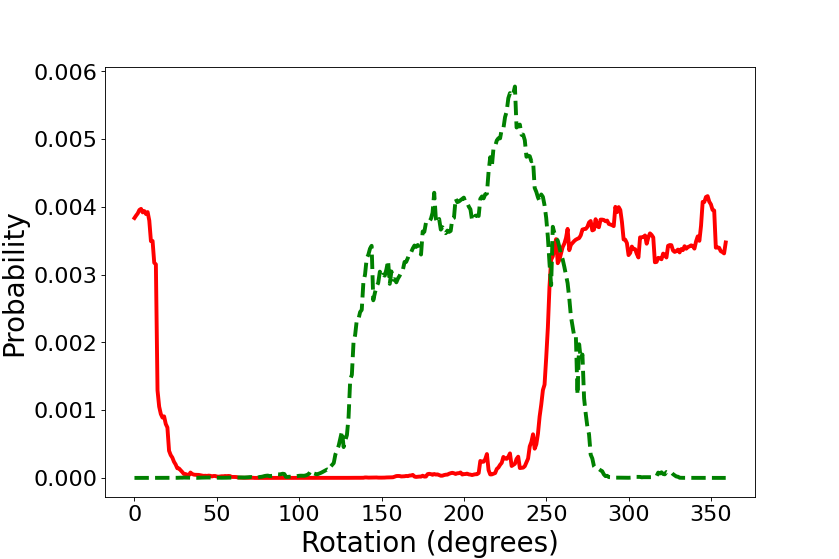}} 
    
   \caption{The pose distribution estimate of our method is shown for both unambiguous and ambiguous views. The top images display the test object in two different poses. Below, the distribution estimates are visualized. The red line indicates the probability of each revolution. The dotted green line shows the same revolution probability for the reflected pose. In the left image, the object's indent is visible, allowing for a single object pose to be identified. In the right image, the indent is not visible, resulting in a much larger pose distribution in both revolution and reflection.}
   % The pose distribution estimate of our method for both unambiguous and ambiguous views. The top images show the test object in two different poses. The distribution estimates are visualized below with the red line indicating the probability of each revolution, and the dotted line indicating the same revolution probability for the reflected pose. In the left image, the indent in the object is visible, and a single object pose is found. However, in the right image, this indent is not visible, and as a result, the pose distribution is much larger, showing uncertainty in reflection.}
   \label{fig:front}
     \vspace{-5mm}
\end{figure}

\section{Introduction}

Recent advances in manufacturing technology have introduced a shift from traditional mass production toward small-batch and one-of-a-kind manufacturing. Additive manufacturing and autonomous robotic assembly are key drivers in this transformation \cite{dalpadulo2022review, prashar2023additive}. Flexible fabrication solutions allow for customization, rapid prototyping, and decentralized production. 
% 
% These technologies are also foundational to the development of autonomous production capabilities in space \cite{zocca2022challenges}. This enables in-space manufacturing without heavy reliance on earth-based supply chains, enabling sustainable long-duration missions. 
% 
Object pose estimation (OPE) plays an important role in flexible robotic assembly, as it enables systems to adapt to novel objects without requiring physical redesign \cite{yokokohji2019assembly}.
The general approach in OPE is a single-pose estimate \cite{hodan2018bop, hodavn2020bop, sundermeyer2023bop, hodan2024bop}. Single-pose estimates align well with many robotic systems, as they enable direct transformation chaining, where components such as camera extrinsics, object poses, and robot TCPs can be linked in a kinematic chain. This enables straightforward computation of grasp poses or end-effector targets using simple matrix multiplications.
However, single-pose estimates are not able to model the aleatoric uncertainty that occurs when visual ambiguity is present. One example of visual ambiguity is seen in many industrial objects, such as idlers, bearings, and screws, which exhibit cylindrical symmetry \cite{yokokohji2019assembly, kleebergerlarge}. Here, the cylindrical shape results in an ambiguity about the correct reflection and revolution of the object. 
For some tasks, pose ambiguity is not relevant to the task's success; however, ignoring pose ambiguity can often lead to failures, potentially damaging the object, equipment, or the robot. To avoid this, extensive checks with additional hardware can be made \cite{hagelskjaer2025good}. But detecting such ambiguities and correcting them is an expensive engineering task that does not align well with flexible systems handling new assemblies with novel objects. 

This limitation of single-pose estimates has led to work on estimating object pose distributions \cite{murphy2021implicit, haugaard2023spyropose, brazi2025corr2distrib}. In these works, the method estimates a probability distribution over poses, formalized as a histogram.
This histogram can then be used to determine the certainty of a given pose. However, all of these methods focus on RGB images for input. 
But, in many industrial scenarios, 3D sensors are used because color information is either unavailable or unreliable. Additionally, by only relying on 3D data the method becomes independent of lighting conditions.
To address this gap, we introduce an object pose distribution estimator that operates solely on 3D data. 

The method is an adaptation of the RGB-based SpyroPose \cite{haugaard2023spyropose} sampling to 3D point clouds. This is achieved by a introducing a feature aggregator that combines spatial and embedding information. To the best of our knowledge, this is the first work to estimate object pose distribution histograms in point clouds using neural networks.

In this paper, we restrict the search space to reflection and revolution. This restriction enables us to generate a complete histogram of rotational uncertainty using our network. Using this, we can define clear rules for interpreting the uncertainty. For example, we may require a single normally distributed estimate, disallow reflection when not physically plausible, or allow revolution symmetry only at specified angular intervals.
This restriction is possible because we rely on 3D data and focus on cylindrical objects. For such objects, most uncertainty can be removed using verification by depth projection \cite{hagelskjaer2024off, hagelskjaer2025good}. However, because of inherent symmetry, depth verification cannot resolve ambiguities in rotation around or reflection across the symmetry axis. As a result, revolution and reflection remain the primary sources of pose uncertainty, and the focus of the uncertainty estimate in this paper.
We demonstrate the effectiveness of this approach in a bin picking scenario. An existing point estimator provides an initial pose, and our method estimates the pose distribution. This enables us to correctly identify ambiguous poses and avoid incorrect grasps, improving reliability and robustness.

The main contributions presented in this paper are:
\begin{itemize}
    % \item A method for object rotation distribution estimation in point clouds
    \item A method for determining revolution and reflection uncertainty in point clouds
    \item A feature aggregator for computing pose likelihoods in point clouds
    \item Verification in a real bin picking scenario
\end{itemize}

We believe this paper will inspire further work in object pose distribution estimation in point clouds.

\section{Related Works}

Visual ambiguities and aleatoric uncertainty are challenging aspects in object pose estimation (OPE). Benchmark datasets typically address this by manually annotating symmetries, thereby requiring only a single-pose estimate for evaluating pose correctness \cite{nguyen2025bop, doumanoglou2016recovering, kleebergerlarge, yang2021robi}.

To mitigate ambiguity in practice, a common strategy in OPE is to generate per-pixel histograms of object correspondences. These 2D–3D correspondences are then processed using PnP-RANSAC \cite{fischler1981random} to compute a final pose. This approach has significantly improved robustness \cite{hodan2020epos, haugaard2022surfemb}, and is currently employed by leading methods in the BOP benchmark \cite{caraffa2024freeze, nguyen2025bop}. However, despite its effectiveness, it remains unclear how to translate per-point distributions into a coherent pose distribution.

\subsection{Distributions in RGB data}
Methods for estimating pose distributions have been actively explored in recent years. For example, \cite{hagelskjaer2019combined} presents an approach where synthetic images are used to estimate the uncertainty of pose estimates. This uncertainty is modeled using a Gaussian distribution and integrated with a simulation of grasping success under pose errors. The resulting framework enables the design of grippers that are robust to OPE uncertainty.

Orientation uncertainty has also been modeled using Bingham and von-Mises distributions \cite{peretroukhin2020smooth, prokudin2018deep}. These parametric models are effective for representing unimodal distributions and have been widely adopted in robotics and computer vision. However, their ability to capture complex uncertainty is limited. Specifically, they cannot represent multimodal uncertainties, such as multiple plausible orientations in visually ambiguous scenes, e.g., when objects exhibit symmetry or are partially occluded.

To address these limitations un-parametric, implicit models have been developed. Using an image crop of the object Ki-Pode \cite{iversen2022ki}, computes heatmaps representing the confidence of each objects keypoint's position. By projecting the object onto these heatmaps an un-normalized likelihood of a given pose can be approximated. 
% marginal distribution of a pose is given. 
Another approach is shown in ImplicitPDF \cite{murphy2021implicit} where an image embedding is combined with a rotation, and then fed to a network which provides an un-normalized likelihood of the rotation.
% By sampling on an equivolumetric partition a pose distribution is obtained.
In both cases normalized using equivolumetric sampling of \textit{SO(3)} is used to provide a normalized likelihood.

% To extend rotation estimation in \textit{SO(3)} to full \textit{SE(3)} 

SpyroPose \cite{haugaard2023spyropose} computes image embeddings which are then extracted using projected keypoints and processed by a Multi Layer Perceptron (MLP). As the keypoints are transformed using a pose this allow for pose distribution in 6 Degrees of Freedom (DoF). However, introducing 6 DoF creates a very large search space. To avoid exhaustive computation, a pyramid approach is used. Starting at an equivolumetric grid at the lowest resolution, followed by importance sampling, this enables SpyroPose to provide pose distributions in \textit{SE(3)}. Similarly to SpyroPose our method extracts embedding in the scene using keypoints. But, as our data is a 3D point cloud nearest neighbor is used instead of the camera projection. In this paper, we do not estimate distributions in \textit{SE(3)}, but only compute uncertainty in revolution and reflection.

A different method for estimating the uncertainty in \textit{SE(3)} is presented in Lie \cite{hsiao2024confronting}. Here a diffusion model is used to represent uncertainty. By using a sampling based strategy the set of poses is recovered.

\begin{figure*}[ht!]
    \vspace{3mm}
    \centering
    \includegraphics[angle=90, width=.95\linewidth]{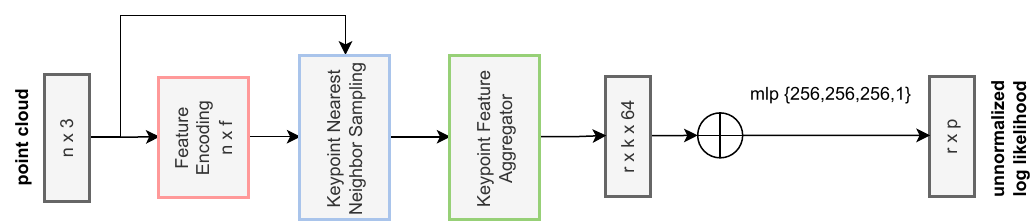}
    \caption{The structure of the developed network. "Feature Encoding" is performed using a PointNet \cite{qi2017pointnet} structure (e.g. DGCNN \cite{dgcnn}, PointNet++ \cite{qi2017pointnet++}). "n" is the size of the input point cloud, "f" is the size of the feature vector from the "Feature Encoding", "r" is the number of sample rotations, and "k" is the number of keypoints.}
    \label{fig:network}
    \vspace{-3mm}
\end{figure*}

\subsection{Distributions in 3D data}
% Similar to RGB based methods with point clouds per-point histograms of object correspondences have been used to improve pose estimation. Here Kabsch-RANSAC \cite{kabsch1976solution,fischler1981random} is generally used to obtain the final pose, but it does not provide pose distributions.   

% For refining pose estimates in point clouds Iterative Closest Point (ICP) \cite{arun1987least} is the standard method. ICP gradually refines the pose by minimizing the distance between the object and scene point clouds. As it is to widespread several different methods have been developed for uncertainty-aware pose estimation using ICP \cite{brossard2020new, censi2007accurate, maken2020estimating, maken2021stein}.

% There is an inherent limitation is the matching of ICP. It is not able to use any learned knowledge of what to focus on when matching. Thus noise can be a limiting factor. Why material is missing. Geometric matching, no learned parameters.

Similar to RGB-based approaches, point cloud methods have employed per-point histograms of object correspondences to enhance OPE. A common technique for computing the pose is Kabsch-RANSAC \cite{kabsch1976solution,fischler1981random}, which robustly estimates transformations but does not yield pose distributions. 

To refine the pose, the standard method is Iterative Closest Point (ICP) \cite{arun1987least}. ICP iteratively minimizes the distance between object and scene point clouds to improve alignment. While ICP itself does not provide uncertainty estimates.  Due to its widespread use, several extensions have been proposed to incorporate uncertainty into ICP-based pose estimation \cite{censi2007accurate, iversen2017prediction, brossard2020new, maken2020estimating, maken2021stein}.
However, ICP has inherent limitations. As it relies solely on geometric matching it does not leverage learned features or contextual knowledge. As a result, it is sensitive to noise and missing data, especially in cases where semantic cues could aid matching. This lack of learned parameters restricts its robustness in complex or ambiguous scenes. This is the same limitation observed with the depth check employed in the initial pose estimation \cite{hagelskjaer2025good, duan2025highprecisionadaptiveselfsupervised}.

% Handling visual ambiguities by point estimates. \cite{hagelskjaer2024keymatchnet}
% Matched using Kabsch-RANSAC \cite{kabsch1976solution,fischler1981random},
% or Voting based \cite{buch2017rotational}.
% \cite{detry2010continuous} Continuous surface-point distributions for 3D object pose estimation and recognition
% \cite{li20203d} 3D object recognition and pose estimation from point cloud using stably observed point pair feature

% One such approach uses 
% Using Bayesian Estimation \cite{maken2020estimating}, 
% Another approach is presented with Stein ICP \cite{maken2021stein}. Here is 

% \newpage

\section{Method}

% The developed method consists of three main parts: a feature encoder, a keypoint feature extractor, and finally, an MLP for scoring. While the structure is based on SpyroPose \cite{haugaard2023spyropose}, several modifications have been made to adapt to 3D data. 

The proposed method comprises three core components: a feature encoder, a keypoint feature extractor, and an MLP for scoring. While the overall architecture is inspired by SpyroPose \cite{haugaard2023spyropose}, several modifications have been introduced to accommodate 3D data.

% Firstly, a PointNet like 
% to compute the and a  neighbor 
% The method consists of three parts. The feature computer, the extractor, and the classifier.
%  "Keypoint Nearest Neighbor Sampling" finds the nearest neighbor for ea 

\subsection{Problem Definition}

Given a set of sample transforms, $T_{sample}$, and a point cloud, compute the likelihood of each sample transform being the correct pose. The set of sample transforms is obtained by combining an initial pose estimate , $T_{init}$, with a set of sample rotations. The sample rotations are used to test for full axial symmetry and reflection. The computation of the sample transforms is shown in Eq.~(1). Where $\theta_{ref}$ is either zero or $180\degree$ and $\theta_{revo}$ goes from zero to $359\degree$.

\begin{equation}
T_{sample} = R_y(\theta_{ref}) R_z(\theta_{revo}) T_{init}
\end{equation}

For each transform, an un-normalized likelihood is computed, and by normalizing across all rotations, the probability of each transform is determined.

%  With one degree resolution this results in 720 sample rotations.
% revolutions and reflections, respectively $R_y(\theta_{reflection})$ and $R_z(\theta_{revolution})$. 

% of each rotation
% As this set of sample rotations does not cover the full 
% We are able to do this because of assuming stuff.

\subsection{Object Keypoints}

Before the network is trained, keypoints are sampled from the CAD model. The keypoints are found similarly to SpyroPose, by using farthest point sampling on the CAD model. In our experiments, we use 32 keypoints, which are sampled only once and remain consistent throughout both the training and test phases.

% To test the different sample transforms. 

% To cover the full set of sample transforms, $T_{sample}$ is applied to the keypoint set. 
%
% To test the rotations 

% The sample transforms, $T_{sample}$, are then applied to 

% % To test the different sample rotations the nearest neighbor of each keypoint is sampled. 
% For each sample rotation the keypoints are then rotated and translated. 
% Using the euclidean distance the nearest neighbor between each keypoint and the scene point cloud is found. 
% 

 % At training and test time the keypoints are transformed using the initial transform and the sample rotations. 

\subsection{Feature Encoding}

Firstly, the scene is encoded using a neural network. We employ a PointNet-like structure, as they have shown very good performance in encoding point cloud data \cite{qi2017pointnet, qi2017pointnet++, dgcnn}. In this paper, we use DGCNN \cite{dgcnn} pre-trained from KeyMatchNet \cite{hagelskjaer2024keymatchnet}, but it could be exchanged with any PointNet-like network \cite{qi2017pointnet, qi2017pointnet++, dgcnn, thomas2019kpconv, zhao2021point}.

The input to the network is a colorless point cloud with six dimensions, consisting of position and normal vectors. The point cloud has a size of 4096. Before processing the point cloud, it is normalized by centering around the object center and scaled by the object radius. For DGCNN, the number of neighbors is set to twenty.

\subsection{Keypoint Nearest Neighbor Sampling}

% When running the network the keypoints are transformed using the 

From the encoded point cloud, the feature points are then sampled by finding the nearest points to the keypoint set. The keypoint set is computed by transforming the keypoints by the sample transforms, $T_{sample}$. Thus, nearest neighbors are found for all the object poses in the sample transform set.
The nearest neighbor is computed using the Euclidean distance between the transformed keypoints and the scene point cloud.

The feature encoding for the nearest point is then used as the feature embedding of the keypoint, $f_j$. As point cloud data also includes spatial information, we incorporate this data as well. Thus, for each keypoint, the spatial difference between the keypoint, $x_i$, and the nearest neighbor, $x_j$, is computed. The embedding and spatial difference are then supplied to the Keypoint Feature Aggregator.

% indices are then found in the input point cloud. 
% Nearest neighbor
% Given the indices the feature embeddings for the keypoints are found. By subtracting the spatial position of the found neighbor the distance for each keypoint is also returned. 
% To determine the correct orientation.
% % Keypoints are extracted from the CAD model using Farthest Point Sampling. For a trained model the same keypoints are always used as the model is trained for the specific positions. 
% To make a prediction for an object pose the keypoints are transformed using the pose estimate. For each keypoint the nearest neighbor in the point cloud is then found. 

% \begin{itemize}
%     \item transform keypoints to position: $x_i$
%     \item compute nearest neighbor for each keypoint $j$
%     \item $x_j$
%     \item $f_j$
% \end{itemize}

\subsection{Keypoint Feature Aggregator}

The keypoint feature aggregator ensures that the spatial information, $x_i-x_j$, is encoded together with the feature encoding, $f_j$. This is achieved by processing both features through an MLP, which results in two 64-dimensional feature vectors, $h_x$ and $h_f$, respectively. These feature vectors are then concatenated and processed by an MLP, $\hat{h}$, returning a size 64 feature vector. The feature computation is shown in Eq.~(2). 

% For both the features and the distance is processed by an MLP. The features are then concatenated and processed by an MLP. 

\begin{equation}
\hat{h}(h_x(x_i-x_j) \oplus h_f(f_j))
\end{equation}

Finally, all keypoint features are concatenated into a single feature vector. This vector is then processed by a three-layer MLP with 256 hidden neurons.

% For the network to make a prediction 
% Transform the keypoints into the scene.
% Nearest neighbor between keypoints.. % We cannot simply project the points
% We add the distance also, instead. 
% Finally all feature are concatenated and processed by a three-layer MLP with 256 hidden neurons as in SpyroPose.
% We compute a feature for the distance and we downsize the object feature to the same size. Concatenate and then 

\subsection{Training}

The network is trained using ADAM optimizer \cite{kingma2014adam} with a learning rate of $10^{-4}$. The network is trained using the InfoNCE loss \cite{oord2018representation} for 4,000 epochs, with 1,000 batches per epoch and a batch size of one. For the three-layer MLP, dropout and point-dropout are both applied. The dropout rate is set to $10\%$.

During training, we utilize data augmentation to enhance generalization. We impose a 1~mm normally distributed error on the object translation, and for the two non-revolutionary axes, a 3-degree variance. On the revolutionary axis, we use a variance of 0.5 degrees. The depth data is augmented according to the strategy in \cite{hagelskjaer2025arrowpose}.

% but we train only using 
% We use the same contrastive loss, but we have a uniform distribution. 
% Number of epochs.
% We train the network for . 

% Data augmentation.

% The data augmentation is similar to the one used in ArrowPose 
% Depth image augmentation.

% Canny edge detection \cite{}

% Point dropout and random movement. 

% Small positional and rotational error.

% \begin{itemize}
%     \item 
% \end{itemize}

\subsection{Inference}

When running the method for inference, the initial pose estimate $T_{init}$ must be provided from an external method. Using the pose estimate, the point cloud is then cropped and normalized. The point cloud and initial pose estimate are then fed to the method, which returns the un-normalized log likelihood of all sample rotations. Finally, a softmax is applied to obtain the pose distribution.

% When computing the pose distribution estimate a 
% Run a pose estimator. Center the pose estimation. Crop the point cloud  and compute the feature encoding. Then apply the sample rotations and compute the un-normalized log likelihood. Finally perform softmax to obtain the pose distribution.

\subsection{Applying the Pose Distribution}

The computed pose distribution is represented as a histogram, indicating the probability that each sampled rotation corresponds to the correct pose. To apply this in practice, it is necessary to define a strategy that determines what level or type of uncertainty is acceptable for a given task.

One straightforward strategy is to require that the pose distribution fit within a single pose. For certain tasks, this constraint is critical to maintaining reliable system performance. For other tasks, the requirement is the absence of uncertainty in reflection, which aligns well with many screw-like objects \cite{yokokohji2019assembly}.
Alternatively, another approach is to accept symmetric rotations, as seen in gears and nuts \cite{yang2021robi}. Depending on the task, these symmetric rotations may be acceptable with or without uncertainty in reflection.
In our experiments, we evaluate both strategies: enforcing a single-pose estimate and ensuring the absence of reflection.
In contrast to the above, a different approach involves determining acceptable uncertainty through simulation \cite{hagelskjaer2019combined, naik2025robotic}. Task execution can be simulated using various grasp poses, allowing the error rate to be integrated with the pose distribution. This provides a probabilistic estimate of task failure given a specific pose distribution.

Handling pose distributions with excessive uncertainty is another important consideration. For some tasks, objects outside the defined threshold can be skipped. For other tasks, a strategy must be specified, such as obtaining new views, reorienting the object, or using physical manipulation to gather new data \cite{hagelskjaer2025good, naik2025robotic}.

% \begin{figure}
% \centering
%     \includegraphics[width=.95\linewidth]{gfx/rotoposenetwork_feature.drawio.pdf}
%     \caption{Keypoint Nearest Neighbor Sampling. First the keypoints are rotated and translated according to the test transform. Nearest neighbor indices are then found in the input point cloud. Given the indices the feature embeddings for the keypoints are found. By subtracting the xyz position of the found neighbor the distance for each keypoint is also returned. }
%     \label{fig:neighborfeat}
%     % \vspace{-6mm}
% \end{figure}

% \newpage

\section{Experiments}

To evaluate the effectiveness of our method, we conduct experiments with two different objects, using both synthetic and real test data. Finally, we implement the method on the real setup and test the grasping performance.

All experiments were performed on a PC environment (Intel Ultra 7 155H CPU and NVIDIA RTX 1000 Ada Generation Laptop GPU). Depth augmentation were performed using OpenCV \cite{opencv_library}, point clouds were processed using Open3D \cite{Zhou2018} and networks were implemented using PyTorch \cite{Paszke_PyTorch_An_Imperative_2019}.

% \subsection{Benchmark}
% We test on the objects of TLESS also shown in SpyroPose. 

\subsection{Objects}

\begin{figure}[t]
% \vspace{3mm}
\centering
    % \subfloat[Object 1.]{\includegraphics[width=.45\linewidth]{gfx/object1.jpg}}\ 
    % \subfloat[Object 2.]{\includegraphics[width=.45\linewidth]{gfx/object2.jpg}}
    \includegraphics[trim={0cm 0.2cm 0cm 0.20cm},clip,width=.99\linewidth]{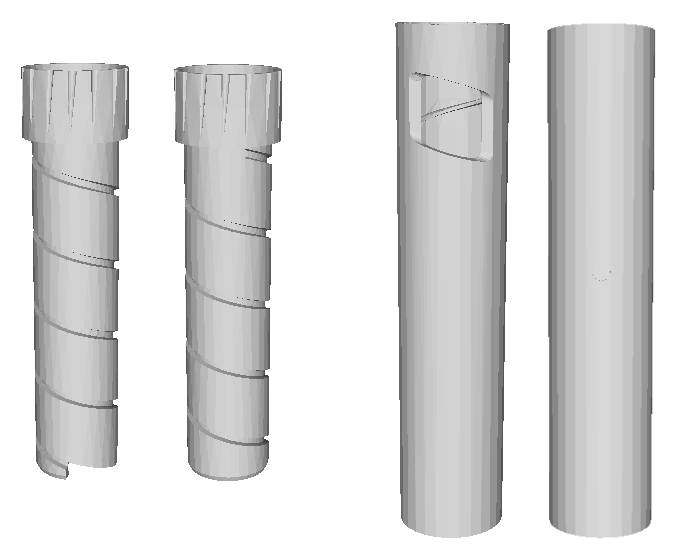}
   \caption{Front and back views of the two objects from our experiments. For Object 1 (left) the correct revolution can only be precisely determined from the indent at the bottom, but the reflection can easily be seen from the screw-head. However, for Object 2 (right) both reflection and revolution can only determined from the square recess. Thus in case of self occlusion the correct reflection is ambiguous.}
   \label{fig:objects}
     % \vspace{-6mm}
\end{figure}

The objects are components of a Novo Nordisk injection device and part of an assembly process, as described in \cite{hagelskjaer2025good, duan2025highprecisionadaptiveselfsupervised}. Both are shown in Fig.~\ref{fig:objects}. Successful insertion requires knowledge of the correct orientation. For Object 1, only the correct reflection is required, while Object 2 also necessitates precise rotational alignment. Object 2 is particularly challenging, as self-occlusion of the square recess can prevent accurate pose estimation.

% Currently the correct 
% % Train on the generated data
% % Get results for pose distribution in each image.
% % Determine the score. Compare with a point estimate.
% % Write something about how mamy operations could be saved.
% To test the 
% To verify the robustness of our method, tests are performed on two different objects. 
% The two objects provide different kinds of visual ambiguity. 
% Self-occlusion for Object 2
% While current point estimates can often find the correct pose, 
% Rotoreflection is only visible 

% Test the correct orientation by single estimate. 

\subsection{Synthetic Data}

\begin{figure}[t]
    \vspace{3mm}
    \centering
    \includegraphics[width=.95\linewidth]{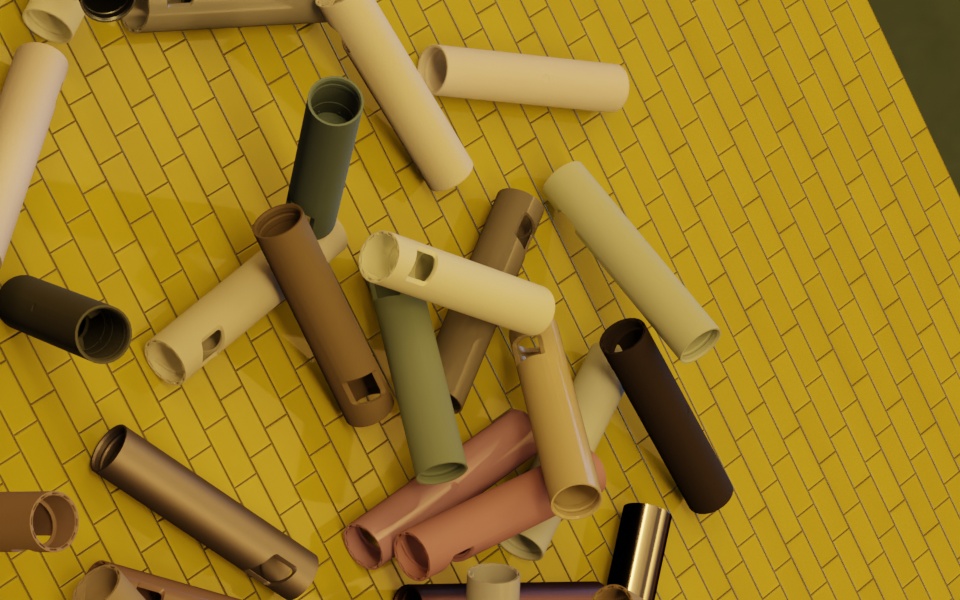}
    \caption{Synthetic image of Object 2.}
    \label{fig:synthimage}
     \vspace{-3mm}
\end{figure}

For experiments with synthetic data, the depth data is generated using BlenderProc \cite{denninger2019blenderproc}. We create 2,000 scenes, of which 100 are retained for testing. Similar to \cite{bregier2017symmetry}, we require that the object have at least $50\%$ visibility, as objects with less visibility generally do not result in stable grasps. This results in 40.459 and 31,622 training instances for objects 1 and 2, respectively, with 2140 and 1743 test instances. An example of the synthetic data is shown in Fig.~\ref{fig:synthimage}.

We test for the ability to correctly classify reflection and the full pose with revolution. The cut-off for accepting a prediction is $99\%$ confidence inside the prediction. For reflection, this is set to the full rotation, and for the full pose, we accept a $15\degree$ range. 

The results of the experiments are shown in Tab.~\ref{tab:testsynth}. From the results, it is seen that our method is able to avoid making predictions in ambiguous scenes and obtain a precision of $100\%$. For the two objects, the coverage, however, is quite different. The reflection is found for almost all cases for Object 1, whereas it is only predicted in $58.9\%$ of cases for Object 2. This is a result of self-occlusion, which means that even with perfect data, the correct orientation is often ambiguous.

\begin{table}[t]
    % \vspace{3mm}
    \centering
    \caption{Results for prediction of reflection and pose for the synthetic test data.}
    \label{tab:testsynth}
    % \begin{tabular}{|c|c|c|c|c|}
    \begin{tabular}{ccccc}
        \hline
        Object               & Instances            & Task             & Coverage ($\%$) & Precision ($\%$) \\ \hline
        \multirow{2}{*}{1}   & \multirow{2}{*}{2140} & Reflection       & 98.8    & 100  \\ 
                             &                      & Pose             & 45.7    & 100  \\ \hline
        \multirow{2}{*}{2}   & \multirow{2}{*}{1743} & Reflection      & 58.9   & 100  \\ 
                             &                      & Pose             & 32.7   & 100  \\ \hline
    \end{tabular}
\end{table}

% Require  for actually be able to grasp the objects. Can easily be verified with the depth data.

% Visibility $50\%$.

% Object 1
% Correct Rotoreflection: 1140/1140/1143
% Correct Orientations: 730/734/1143
% Object 2
% Correct Rotoreflection: 617/617/850
% Correct Orientations: 407/407/850

% And then with noise...

% Correct Rotoreflection: 2121/2122/2140
% Correct Orientations: 0.0/0/2140
% Correct Rotoreflection: 1057/1057/1743
% Correct Orientations: 364/364/1743

\subsection{Bin Picking Data}

\begin{figure}[t]
    \vspace{3mm}
    \centering
    \includegraphics[angle=0, width=.98\linewidth]{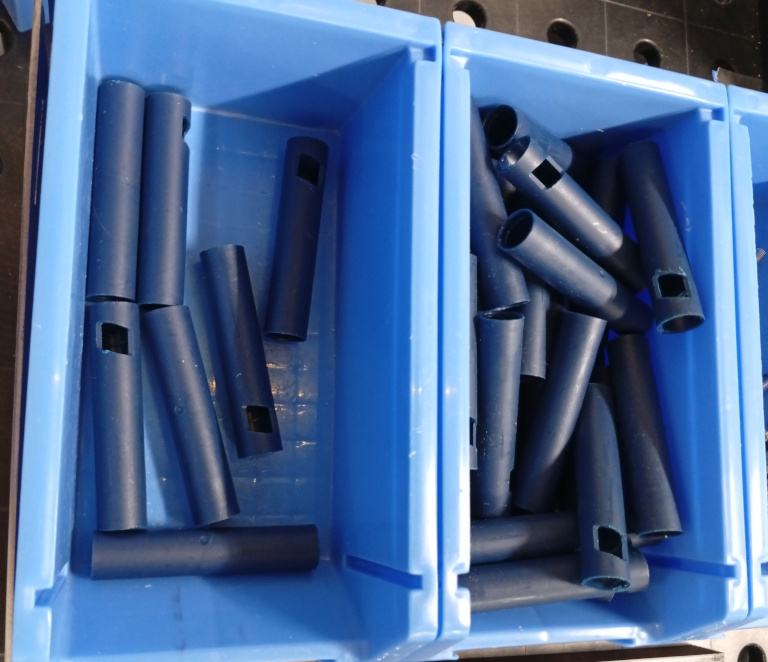}
    \caption{Example of Object 2 in the bins.}
    \label{fig:bin}
    % \vspace{-6mm}
\end{figure}

In the existing set-up pose estimates were obtained using KeyMatchNet \cite{hagelskjaer2024keymatchnet}, and incorrect poses were filtered out using a depth check with a threshold of $2.5,\mathrm{mm}$. An example of the bins is shown in Fig.~\ref{fig:bin}.
However, in cases of visual ambiguity, single-pose estimates do not convey the pose uncertainty. To ensure accurate pose estimation, the system performs an in-hand pose check. This procedure is time-consuming, and for Object 2, it still results in ambiguous reflection estimation. Consequently, an additional fixture check was implemented \cite{duan2025highprecisionadaptiveselfsupervised}.
The pose estimation from the system resulted in incorrect reflection estimates of $86.4\%$ and $53.8\%$ for Object 1 and Object 2, respectively.

\begin{figure}[t]
    % \vspace{3mm}
    \centering
    \subfloat{\includegraphics[trim={5.5cm 0.5cm 0.5cm 2.5cm },clip, width=.48\linewidth]{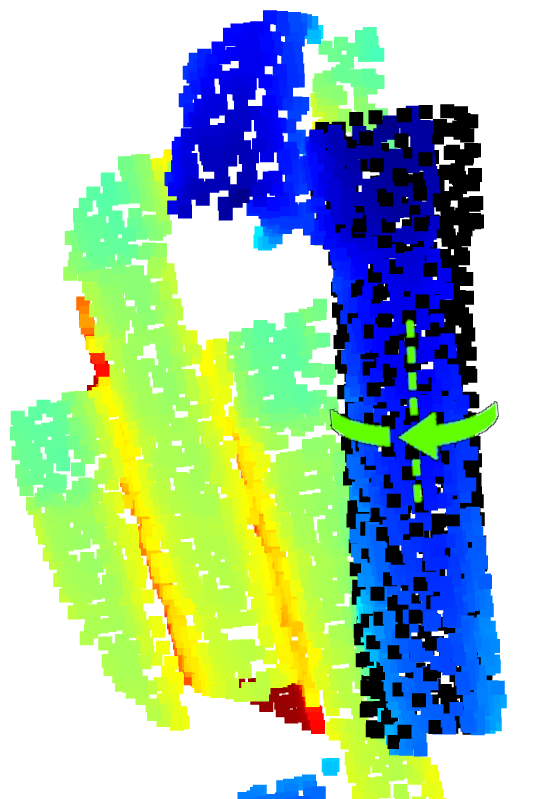}}\
    \subfloat{\includegraphics[trim={3.5cm 2.5cm 3.5cm 2.5cm},clip, width=.48\linewidth]{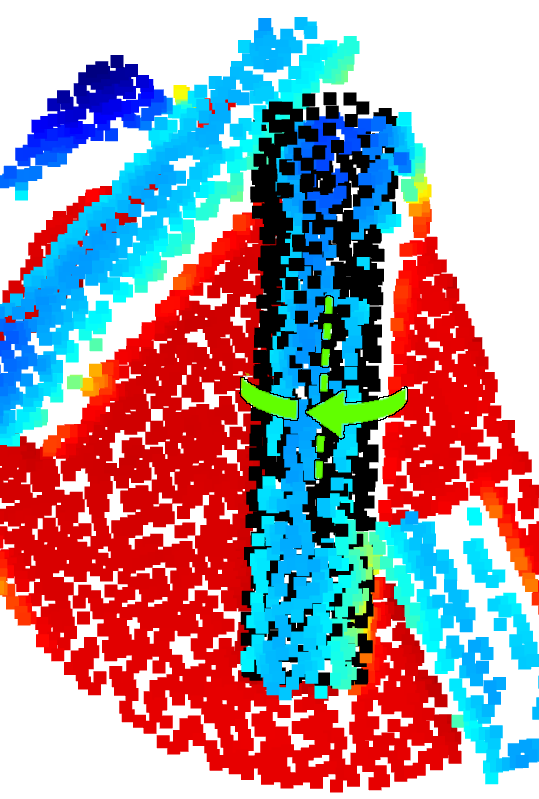}} 
    % \subfloat[Unambiguous view with a single object pose.]{\includegraphics[width=.48\linewidth]{gfx/Screenshot from 2025-07-29 13-51-29.png}}\
    % \subfloat[Ambiguous view with unknown revolution and rotoreflection.]{\includegraphics[width=.48\linewidth]{gfx/Screenshot from 2025-07-29 13-52-24.png}} 
    
     \vspace{-1.5mm}
    
    \subfloat[View where the tack at the bottom is visible. Resulting in a single prediction.]{\includegraphics[trim={0cm 0 1.5cm 1cm},clip, width=.48\linewidth]{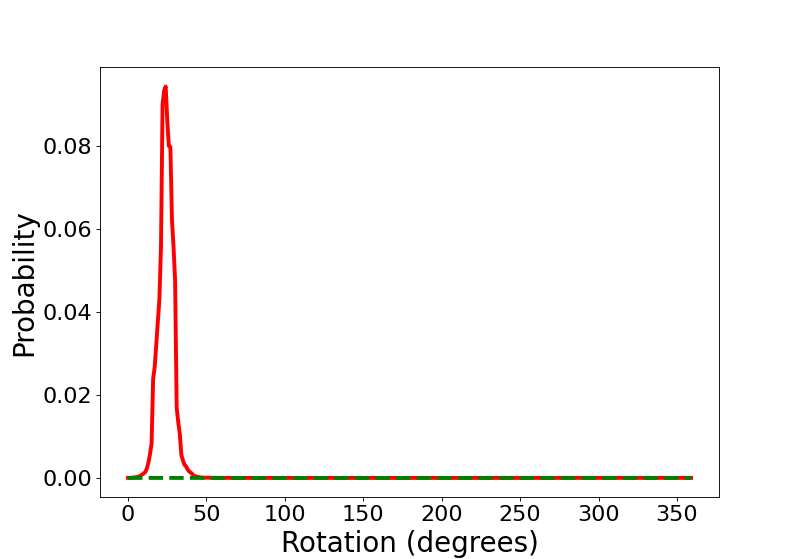}}\
    \subfloat[View without the tack at the bottom visible. Resulting in a larger uncertainty in pose.]{\includegraphics[trim={0cm 0 1.5cm 1cm},clip,width=.48\linewidth]{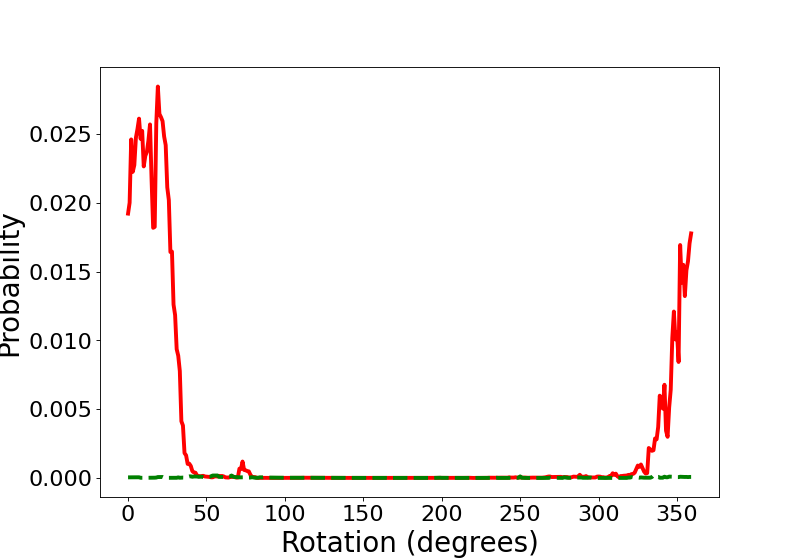}}

   \caption{The pose distribution estimate of our method visualized, for two instances of Object 1. When the tack at the bottom is not visible, the pose distribution becomes larger.}
   \label{fig:object1}
     % \vspace{-3mm}
\end{figure}

We test our method using the real data collected from the set-up \cite{hagelskjaer2025good}. The training data is the same synthetic data from the previous section. The result of our method on the real data is presented in Table~\ref{tab:testreal}. For all objects in the test set, the precision is $100\%$. However, task coverage varies significantly. For Object 1, reflection prediction coverage is $87\%$, which is substantially higher than the $4.5\%$ coverage for full pose prediction. This discrepancy likely stems from noise in the real data. As the grooves on the object are not visible, only the bottom indent can be used to determine revolution. Due to noise interference, only a few samples achieve sufficient confidence. Real test data for Object 1 is shown in Fig.~\ref{fig:object1}.
For Object 2, the coverage of both tasks is more balanced, as the square recess must be visible to determine either. Although the coverage is lower, it remains within a usable range.

With $100\%$ precision across all tasks and objects, the system can operate autonomously without requiring additional engineering solutions such as a second camera or mechanical verification. This enables flexible production setups where new objects, such as those from additive manufacturing, can be integrated remotely.
The overall recall for correct reflection estimation, disregarding distribution requirements, is $98\%$ for Object 1 and $80\%$ for Object 2. These results significantly outperform current systems, suggesting that our approach could enhance existing setups with engineered checks.

\begin{table}[t]
    \vspace{3mm}
    \centering
    \caption{Results for prediction of reflection and pose for the real test data.}
    \label{tab:testreal}
    \begin{tabular}{ccccc}
        \hline
        Object               & Instances            & Task             & Coverage ($\%$) & Precision ($\%$) \\ \hline
        \multirow{2}{*}{1}   & \multirow{2}{*}{200} & Reflection       &  87     & 100  \\ 
                             &                      & Pose             & 4.5    & 100  \\ \hline
        \multirow{2}{*}{2}   & \multirow{2}{*}{200} & Reflection       &  43     & 100  \\ 
                             &                      & Pose             &  31     & 100  \\ \hline
    \end{tabular}
\end{table}

% \begin{figure}[t]
%     \vspace{3mm}
%     \centering
%     \subfloat{\includegraphics[trim={5.5cm 0.5cm 0.5cm 2.5cm },clip, width=.48\linewidth]{gfx/object1_1.png}}\
%     \subfloat{\includegraphics[trim={3.5cm 2.5cm 3.5cm 2.5cm},clip, width=.48\linewidth]{gfx/object1_2.png}} 
%     % \subfloat[Unambiguous view with a single object pose.]{\includegraphics[width=.48\linewidth]{gfx/Screenshot from 2025-07-29 13-51-29.png}}\
%     % \subfloat[Ambiguous view with unknown revolution and rotoreflection.]{\includegraphics[width=.48\linewidth]{gfx/Screenshot from 2025-07-29 13-52-24.png}} 
    
%      \vspace{-2mm}
    
%     \subfloat{\includegraphics[trim={0cm 0 1.5cm 0},clip, width=.48\linewidth]{gfx/Figure_Obj_1_28.png}}\
%     \subfloat{\includegraphics[trim={0cm 0 1.5cm 0},clip,width=.48\linewidth]{gfx/Figure_Obj_1_19.png}} 

%    \caption{The pose distribution estimate of our method visualized, for two instances of Object 1. When the tack at the bottom is not visible, the pose distribution becomes larger.}
%    \label{fig:object1}
%      % \vspace{-3mm}
% \end{figure}

%%%%%%%%%%%%%%%%%%% NEW STUFF HAS BEEN ADDED %%%%%%%%%%%%%%%%%%%%%%%%%%%%%%%%%%%%

\subsection{Bin Picking Test}

\begin{figure}[t]
    \vspace{3mm}
    \centering
    \includegraphics[angle=0, width=.75\linewidth]{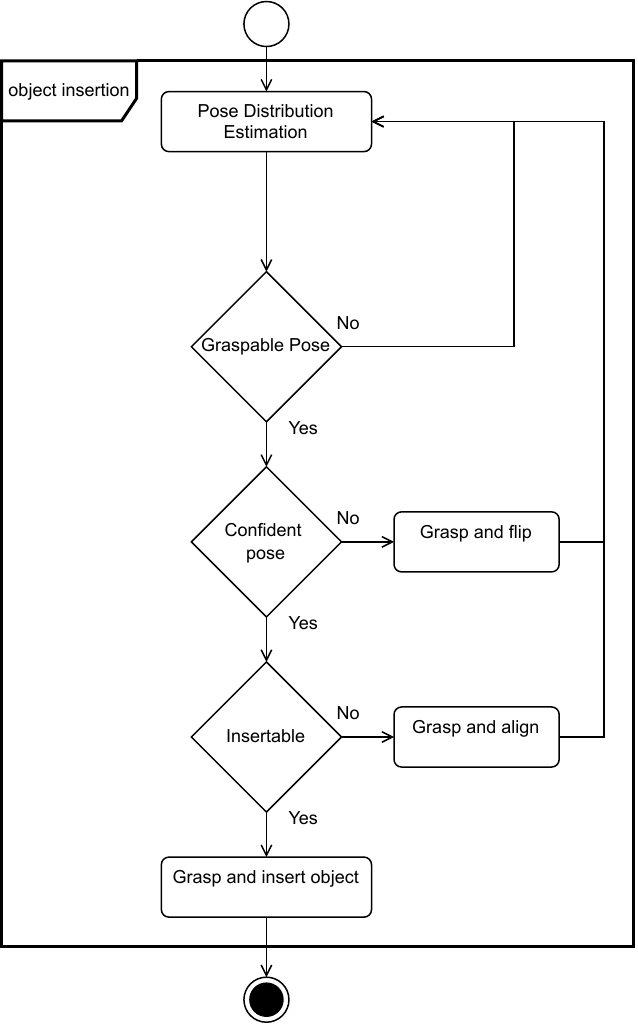}
    \caption{Activity diagram of the bin picking process using confidence intervals. If no object poses with high confidence are available the robot grasps an object and reorients it.}
    \label{fig:strat}
    \vspace{-6mm}
\end{figure}

To test the effectiveness of the developed method in a bin picking scenario, we implemented the algorithm on a real setup. We compare our method with the existing system, which relies on a second camera for in-hand pose estimation and a fixture for verification. As our method returns the complete pose distribution, we do not rely on either of these additional steps. However, if all object poses are uncertain, the system cannot perform any grasps. To ensure that all objects can be grasped and inserted, we develop a re-orientation strategy. First, all object poses are found, and if none of the poses are certain, an object is grasped and flipped into a random position. If a pose is certain, but the resulting grasp is not usable for insertion, the object is aligned so that it can be inserted in the next grasp. Finally, if the pose is confident and the object can be inserted, this is simply performed. This bin picking strategy is shown in Fig.~\ref{fig:strat}.

For both methods, we operated the system until 10 successful insertions were completed. The results appear in Tab.~\ref{tab:testbinpick}. From these results, several key observations emerge. First, both methods achieved insertions without critical failures. Our method required 20 grasps to accomplish 10 insertions, resulting from five rotation adjustments and five object flips. This slightly exceeds the 18 grasps required by the original setup. Crucially, our system did not produce any incorrect insertions, indicating that supplemental hardware and software components are unnecessary. In two alignments, the resulting pose was not suitable for insertion. System performance could be enhanced by adopting improved alignment strategies.

% However, when comparing the number of incorrect poses that were found and then grasped, the methods differ quite drastically.  

\begin{table}[t]
    % \vspace{3mm}
    \centering
    \caption{Results for bin picking of 10 objects. Comparison with the original setup.}
    \label{tab:testbinpick}
\begin{tabular}{lccc}
Method                  & Grasps         & Incorrect Insertions & Failures \\ \hline
Previous                & 18             & 8                  & 0 \\ \hline
Ours                    & 20             & 0                   & 0    
\end{tabular}
    \vspace{-3mm}
\end{table}

\begin{figure}[t]
    \vspace{3mm}
    \centering
    \includegraphics[angle=0, width=.98\linewidth]{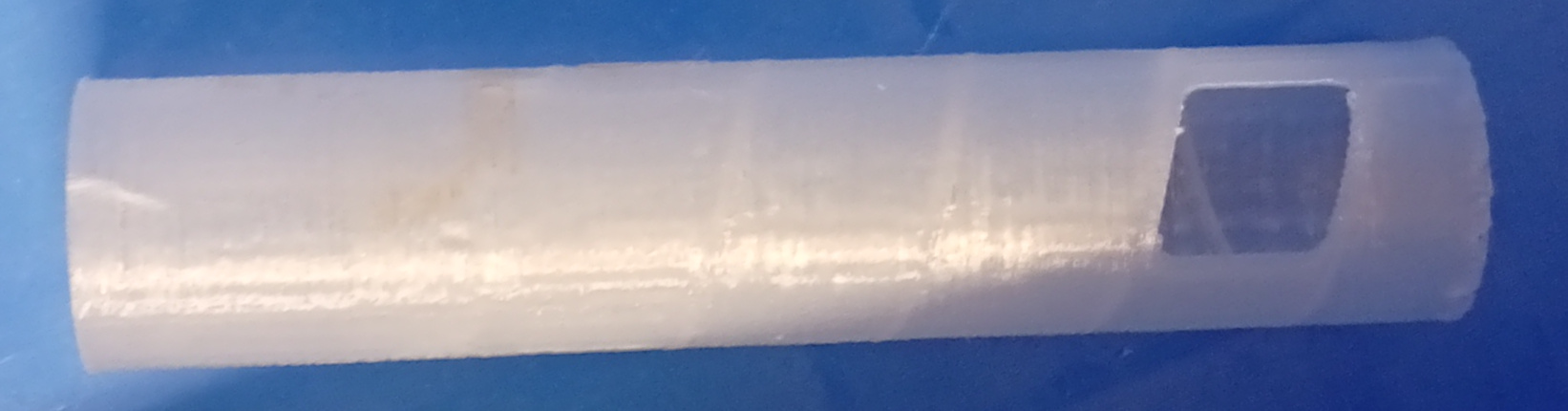}
    \caption{Example of the 3D printed test object.}
    \label{fig:additive}
    % \vspace{-6mm}
\end{figure}

\subsection{Additive Manufacturing}

% We deployed the algorithm on the existing setup to evaluate its grasping performance. The task required grasping Object 2 and inserting it into a fixture. We omitted in-hand pose estimation as the pose distribution eliminates ambiguous poses.

To assess the algorithm's effectiveness in an additive manufacturing context, we used a 3D-printed version of the object made from a semi-transparent material. An example of the printed test object is shown in Fig.~\ref{fig:additive}. The test was performed using the grasping strategy introduced above. The system executed ten successful insertions without any errors. All ambiguous poses were consistently reoriented until insertions could be performed. 

% However, it was observed that the system exhibited low confidence in certain views, despite the square recess being clearly visible.

\subsection{Ablation Study}

To test the contribution of the different parts of the feature aggregator, an ablation study is performed. We train a version of the model excluding either the spatial information or the feature encoding. We test the network's performance on the synthetic test data for object 2. The results are shown in Tab.~\ref{tab:abltestsynth}. From the results, it is evident that combining the features yields better performance than using the feature encoding alone. It is also observed that, by using only spatial information, the network is unable to learn a representation of pose uncertainty. 

\begin{table}[t]
    % \vspace{3mm}
    \centering
    \caption{Ablation study results for prediction of reflection and pose for object 2 on the synthetic test data.}
    \label{tab:abltestsynth}
    % \begin{tabular}{|c|c|c|c|c|}
    \begin{tabular}{ccccc}
        \hline
        Omitted               & Instances            & Task             & Coverage ($\%$) & Precision ($\%$) \\ \hline
        \multirow{2}{*}{None}   & \multirow{2}{*}{1743} & Reflection      & 58.9   & 100  \\ 
                             &                      & Pose             & 32.7   & 100  \\ \hline
        \multirow{2}{*}{Spat.}   & \multirow{2}{*}{1743} & Reflection      & 56.0   & 100  \\ 
                             &                      & Pose             & 29.9   & 99.8  \\ \hline
        \multirow{2}{*}{Feat.}   & \multirow{2}{*}{1743} & Reflection      & 0   & N/A  \\ 
                             &                      & Pose             & 0   &  N/A \\ \hline
    \end{tabular}
    \vspace{-3mm}
\end{table}

\subsection{Run-time}

\begin{table}[t]
    \vspace{3mm}
    \centering
    \caption{The run-time of a single pass of the algorithm.}
    \label{tab:cum_runtime}
    % \begin{tabular}{|l|c|c|}
    \begin{tabular}{lcc}
        \hline
        Task & Time [ms] & Cum. Time [ms] \\ \hline
        Loading point cloud &  25.5   & 25.5 \\
        Feature Encoding    & 4.9     & 30.4 \\
        Nearest Neighbor    & 0.2     & 30.6 \\
        Feature Aggregator  & 0.7     & 31.3 \\
        Model Head          & 1.3     & 32.5 \\
        \hline
    \end{tabular}
     \vspace{-3mm}
\end{table}

Another aspect of the algorithm's usability is its run-time. The run-time of each part of the algorithm is shown in Tab.~\ref{tab:cum_runtime}. The cumulative run-time for a single object is 32.5 milliseconds. This would enable near real-time processing of objects during the robot's operation.

An important factor for further development is the run-time for the pose sampling. If the method were extended to cover \textit{SE(3)}, this part would increase, whereas the run-time for loading point clouds and feature encoding would remain static. The cumulative time for sampling the 720 poses is 2.1 milliseconds. This is a small part of the full run-time and would allow the algorithm to be used for estimating uncertainty in \textit{SE(3)}.

% \subsection{Bin picking benchmark dataset}

% Applied on top of ArrowPose. And without ArrowPose.

% % \subsection{Ablation Study}
% % How about simply comparing with using the matches of the RANSAC.
% % Not including the 
% % Different backends.

% \begin{figure}[t]
% \centering
%     \subfloat[Unambiguous view.]{\includegraphics[width=.48\linewidth]{gfx/unp1.png}}\ 
%     \subfloat[Resulting certain pose estimate.]{\includegraphics[width=.48\linewidth]{gfx/und1.png}}

%     \subfloat[Ambiguous view.]{\includegraphics[width=.48\linewidth]{gfx/unp2.png}}\ 
%     \subfloat[Resulting uncertain pose estimate.]{\includegraphics[width=.48\linewidth]{gfx/und2.png}}
    
%    \caption{The uncertainties estimated by our method.}
%    \label{fig:uncertainty1}
%      % \vspace{-6mm}
% \end{figure}

\section{Conclusion and Future Work}
In this paper, we present a method for estimating object pose distributions in 3D point clouds. Our method estimates uncertainty in revolution and reflection. Tests were conducted on two objects from a bin picking scenario. The method was also implemented on a real system and used for grasping. Experiments demonstrate that the method can model the uncertainty arising from visual ambiguities. Thus, our method does not make any incorrect pose predictions and can be used reliably in uncertain environments. This will enable more robust and flexible robotic applications that can still operate correctly in the presence of visual ambiguity.

In further work, the method can be extended to \textit{SE(3)} distributions to cover full 6 DoF pose uncertainty estimation. This will require sampling and training similar to SpyroPose \cite{haugaard2023spyropose}, but the network structure would not have to be changed. Additionally, the method could be tested on benchmark datasets and potentially compared with RGB-based methods.
Another topic for future research is to evaluate the impact of different feature encoders. We have used the well-known DGCNN \cite{dgcnn}, but several others could be used. Color information could also be incorporated to test the effectiveness of a multi-modal encoder.

% In this paper, we present a method for estimating object pose distributions in 3D point clouds. Our method estimates uncertainty in revolution and reflection. Tests were performed on two objects from a real bin picking scenario. Experiments demonstrate that the method can effectively avoid making predictions when visual ambiguity is present. And thus, it does not make any incorrect predictions.

% In further work, the method can be extended to \textit{SE(3)} distributions to cover full 6 DoF pose uncertainty estimation. This will require sampling and training similar to SpyroPose \cite{haugaard2023spyropose}, but the network structure would not have to be changed. The method could also be tested on benchmark datasets and potentially be compared with RGB-based methods.

% Another topic for future work is evaluating the impact of different feature encoders; we have used the well-known DGCNN \cite{dgcnn}, but several others could be implemented. Color information could also be incorporated to test the effectiveness of a multi-modal encoder.

\section*{Acknowledgment}
The authors gratefully acknowledge the helpful discussions and feedback provided by Rasmus Laurvig Haugaard.
% The authors thank Rasmus Laurvig Haugaard for helpful discussions and valuable feedback.

% \bibliographystyle{IEEEtran} % We choose the "plain" reference style
% \bibliography{egbib}

% {\small
% \bibliographystyle{ieee_fullname}
% \bibliography{egbib}
% }
% \IEEEtriggeratref{20}
% \enlargethispage{+20mm}
\bibliographystyle{IEEEtran}
\bibliography{egbib}
\clearpage
\end{document}